\documentclass[10pt,twocolumn,letterpaper]{article}

\usepackage[pagenumbers]{cvpr} 
\usepackage[linesnumbered,ruled,vlined]{algorithm2e}
\usepackage{multirow}
\usepackage{placeins}
\usepackage{float}
\usepackage{dblfloatfix}
\definecolor{cvprblue}{rgb}{0.21,0.49,0.74}
\usepackage[pagebackref,breaklinks,colorlinks,allcolors=cvprblue]{hyperref}

\def\paperID{43389} 
\def\confName{CVPR}
\def\confYear{2026}

\title{Generative Adversarial Perturbations with Cross-paradigm \\ Transferability on Localized Crowd Counting}

\author{
Alabi Mehzabin Anisha \quad Guangjing Wang \quad Sriram Chellappan\\
University of South Florida, Tampa, Florida, USA\\
{\tt\small \{aanisha, guangjingwang, sriramc\}@usf.edu}
}
\begin{document}
\maketitle
State-of-the-art crowd counting and localization are primarily modeled using two paradigms: density maps and point regression. Given the field's security ramifications, there is active interest in model robustness against adversarial attacks. Recent studies have demonstrated transferability across density-map-based approaches via adversarial patches, but cross-paradigm attacks (i.e., across both  density map-based models and point regression-based models) remain unexplored. We introduce a novel adversarial framework that compromises both density map and point regression architectural paradigms through a comprehensive multi-task loss optimization. For point-regression models, we employ scene-density-specific high-confidence logit suppression; for density-map approaches, we use peak-targeted density map suppression. Both are combined with model-agnostic perceptual constraints to ensure that perturbations are effective and imperceptible to the human eye. Extensive experiments demonstrate the effectiveness of our attack, achieving on average a $7\times$ increase in Mean Absolute Error compared to clean images while maintaining competitive visual quality, and successfully transferring across seven state-of-the-art crowd models with transfer ratios ranging from $0.55$ to $1.69$. Our approach strikes a balance between attack effectiveness and imperceptibility compared to state-of-the-art transferable attack strategies. 
The source code is available at \url{https://github.com/simurgh7/CrowdGen}
\section{Introduction} \label{sec:intro}
Localized crowd counting~\cite{zhang2016single,idrees2018composition, wang2020nwpu,sindagi2020jhu,song2021choose,liang2022focal, du2023redesigning,lin2024gramformer,song2021rethinking,  liu2023point, chen2024improving, liu2019recurrent, sam2020locate, anisha2025scale} extends beyond population estimation to precisely pinpointing individual locations within crowded scenes. This capability is crucial for many applications like public safety management, retail analytics and epidemiological tracking. The research community has developed two predominant architectural paradigms for this task: density map-based approaches~\cite{song2021choose,liang2022focal, du2023redesigning,lin2024gramformer} that regress spatial density distributions followed by post-processing to locate individuals, and point regression methods that directly output coordinates and confidence scores in an end-to-end manner \cite{song2021rethinking,  liu2023point, chen2024improving}. Despite their widespread adoption, the security and robustness of these deployed models remain largely unexamined, posing significant risks in real-world scenarios where malicious attacks could undermine safety protocols, operational planning, and policy making.
 \begin{figure}
    \centering
    \includegraphics[width=0.45\linewidth]{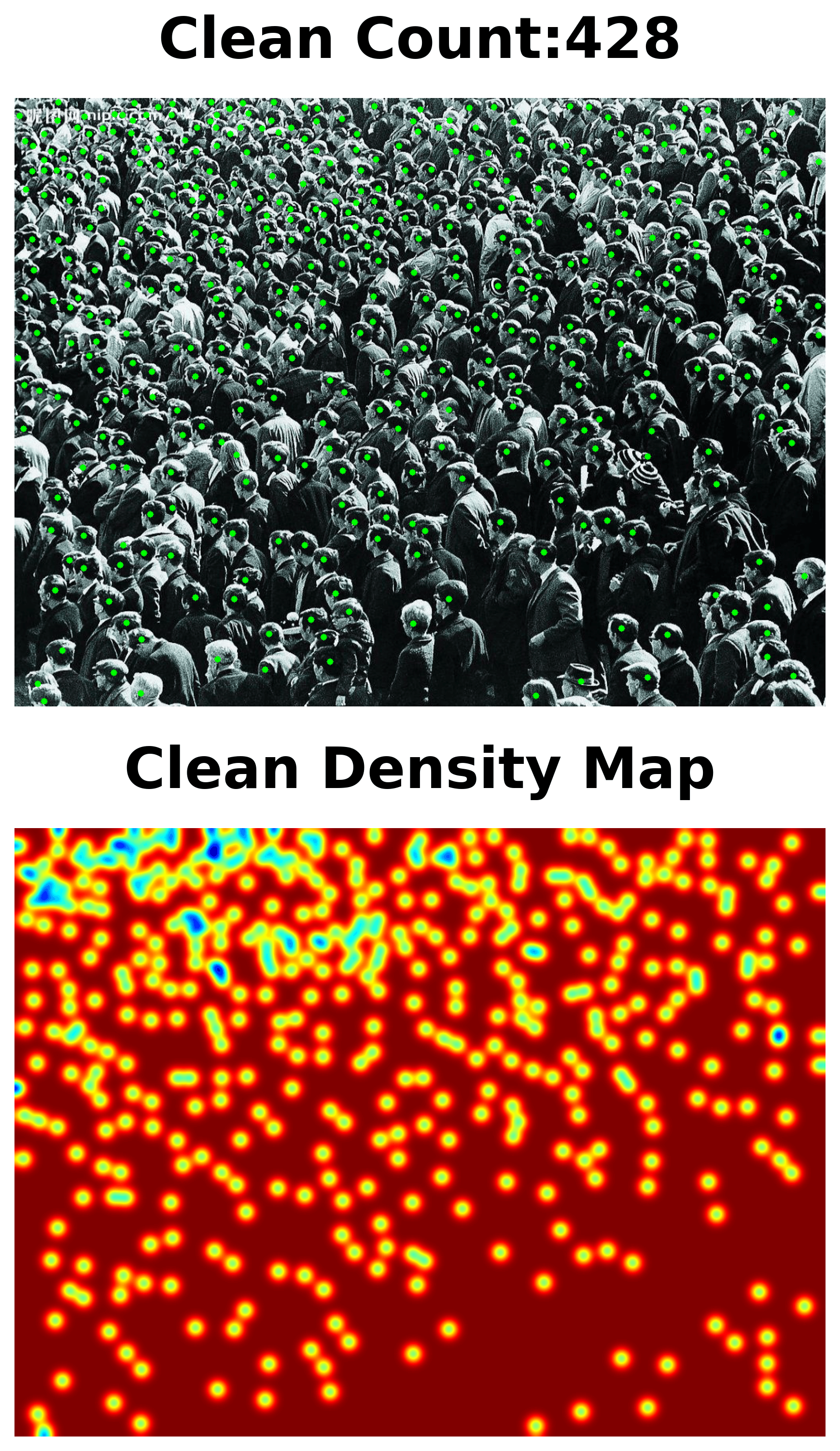}
    \includegraphics[width=0.45\linewidth]{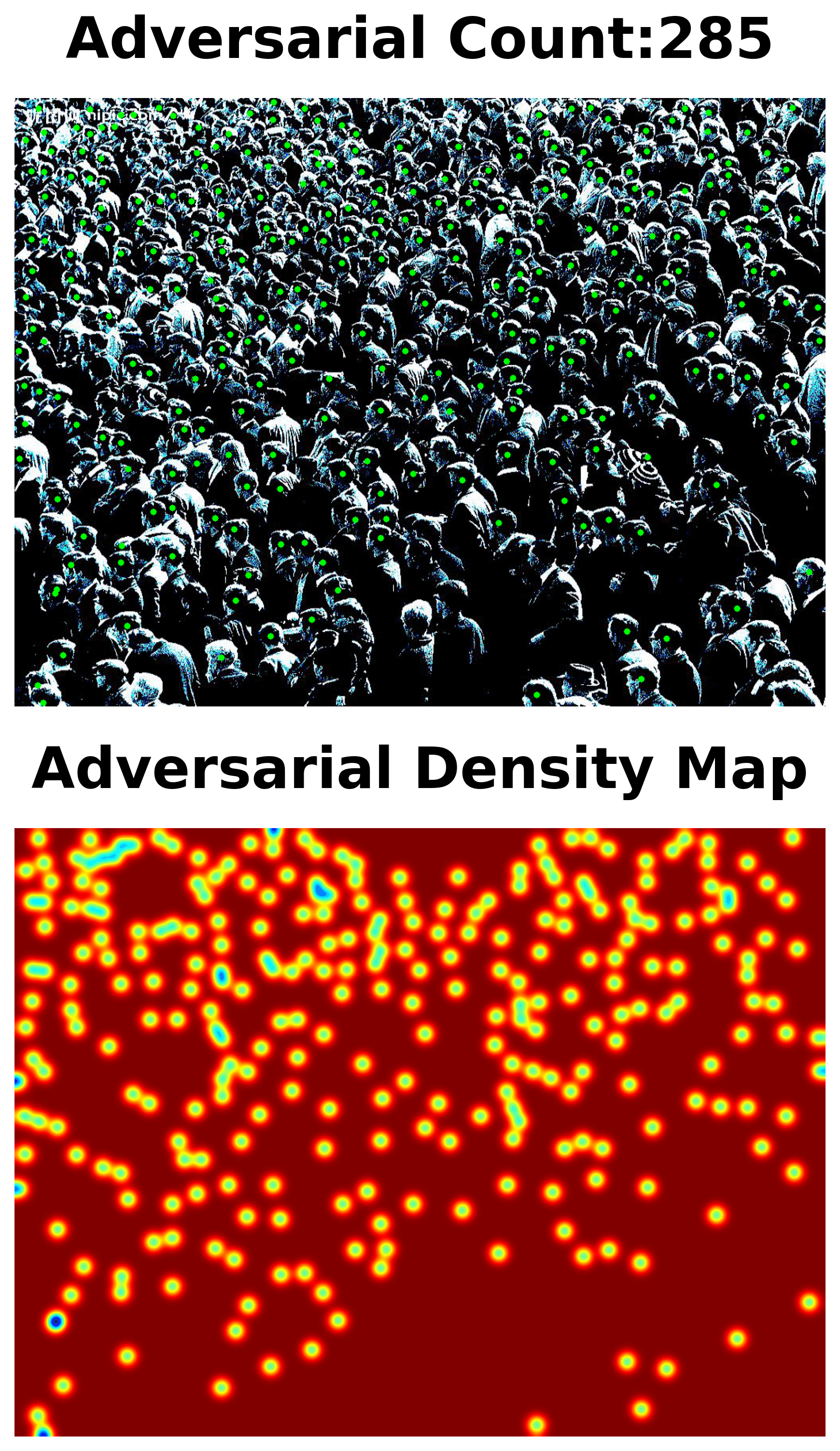}
    \vspace{-2mm}
    \caption{Localized crowded counting predictions and density maps in clean and adversarial images designed in our work.}
    \label{fig:comparision}
\end{figure}
\noindent Crowd counting models inherit the vulnerability of neural networks to adversarial attacks~\cite{biggio2013evasion}. Recent efforts in attacking crowd counting models, such as APAM~\cite{wu2021towards} and PAP attack models~\cite{liu2022harnessing}, rely on generating patches to attack the model's outputs. However, existing work suffers from critical limitations of striking the right balance in achieving stealth, measured by Peak Signal-to-Noise Ratio (PSNR), and compromising counting performance, measured by Mean Absolute Error (MAE). More fundamentally, current approaches, exist in two extremes: either achieving high stealth with higher PSNR, but low MAE (e.g., PAP~\cite{liu2022harnessing}, GE-AdvGAN~\cite{zhu2024ge} with $\text{PSNR} \geq 22 \text{dB}$ but $\text{MAE} \leq 120$), or achieving high MAE in crowd counting, but produce visually perceptible attack outcomes with low PSNR (e.g., DiffAttack~\cite{chen2024diffusion} with $\text{PSNR} = 11 \text{dB}$). 

\noindent In addition, localized crowd-counting systems are generally black-box, meaning that attackers have no knowledge of the model architecture, parameters, or specifications. The newly emerging designs of transfer-based black-box attacks represent a middle ground, where attackers generate adversarial examples using surrogate models without requiring access to the target systems \cite{xie2019improving,wang2021admix,zhu2024learning,zhu2023boosting,wang2021feature,xiao2018generating,xiang2022egm, chen2024diffusion,zhu2024ge}. The transferability
amplifies attack impact in black-box and real-world conditions~\cite{wang2023beyond}. However, existing transferability studies of APAM~\cite{wu2021towards} and PAP~\cite{liu2022harnessing} attack models have been limited to single-paradigm crowded-counting models such as density map-based methods, without consideration of transferability to point regression-based methods. 

\noindent In this work, we propose a simple but effective adversarial attack framework that achieves  high attack performance, stealth and transferability. The proposed framework is capable of attacking diverse crowd counting models without requiring adaptations specific to each model. For example, in~\figurename~\ref{fig:comparision}, we demonstrate that our proposed attack, designed based on the density map-based model SASNet~\cite{song2021choose}, substantially degrades the counting accuracy of another point-based crowd-counting model, P2PNet~\cite{song2021rethinking}, with 20.68 PSNR. 

Our attack  model leverages a lightweight generative UNet model and novel loss functions to optimize perturbations that are both effective and visually subtle. Developing black-box adversarial attacks for localized crowd counting involves two key challenges: (i) Cross-paradigm transferability, which refers to creating attacks that remain effective across fundamentally different model architectures, such as density map-based and point regression methods; (ii) Attack versus stealth trade-off, which requires generating perturbations that degrade counting performance while remaining visually imperceptible.

In summary, our core contributions are listed as follows:
\begin{itemize}
\item We proposed the first framework for cross-paradigm adversarial attacks on localized crowd counting models, demonstrating successful transferability between density map and point regression architectures, achieving as high as a transfer ratio (TR) $= 1.69$. 
\item We designed a multi-task loss to optimize effective, yet imperceptible perturbations, to achieve a superior attack-stealth trade-off compared to existing methods. Specifically, for density-map approaches, we design density suppression loss, and for point-regression, we design logit suppression loss to maximize the attack strength. 
\item We conduct experiments to evaluate the transferability and efficiency of the proposed attack. For density-map models, for example, our attack in the SHHA dataset~\cite{zhang2016single} results in an average $9\times$ MAE increase and with a $50\%$ miss rate. 
For point-regression based methods, a $9.67 \times$ MAE increase and $47.72\%$ miss rate with the same PSNR and SSIM levels is achieved in our attack.
\end{itemize}

\section{Related Work}\label{sec:rel}
\subsection{Localized Crowd Counting}
The modeling process for crowd localization and counting mainly includes two design paradigms: density-map and point-regression-based methods. The other approaches, such as object detection-based, are less adopted due to computationally inefficient training and complex post-processing requirements like non-maximum suppression in dense crowds~\cite{liu2019point, sam2020locate}. Density-map-based methods leverage multi-scale representations to regress spatial density distributions from input images, followed by non-differentiable post-processing to extract individual locations and counts~\cite{song2021choose,liang2022focal, du2023redesigning,lin2024gramformer}. In contrast, point-regression methods operate end-to-end, predicting coordinate points and confidence scores without intermediate density estimation or complex post-processing~\cite{song2021rethinking, liu2023point, chen2024improving}. 

\noindent Despite their architectural differences, both paradigms remain vulnerable to adversarial attacks. Recent work has explored transferable adversarial example generation using patches, with PAP~\cite{liu2022harnessing} and APAM~\cite{wu2021towards} demonstrating vulnerabilities in density-based crowd counting models. However, these approaches remain limited to a single paradigm, leaving cross-paradigm security analysis unexplored. Crucially, they provide no analysis of why subtle, perturbation-based attacks are uniquely challenging in crowd scenes, a gap our work directly addresses.
\subsection{Transferability of Adversarial Perturbations}
White-box approaches hold the assumption of access to the model architecture and weights \cite{goodfellow2014explaining,moosavi2016deepfool,madry2017towards, carlini2017towards, wang2023beyond}, while black-box approaches rely on the ability to perform queries and process responses for attack \cite{chen2017zoo,athalye2018obfuscated,brendel2017decision,chen2020hopskipjumpattack, guo2024wavepurifier, zhou2024optical, guo2023phantomsound}. In addition, the surrogate model-based methodology is also widely adopted for black-box attacks.
Attackers can generate adversarial examples using a surrogate model in a white-box manner to attack a black-box model. The current landscape of optimizing adversarial attack transferability falls into two major categories: (i) optimization-based, and (ii) generation-based. 

\noindent Optimization-based methods iteratively optimize a loss function $\mathcal{L}$ on the gradients from a surrogate model $f_s$ to generate adversarial perturbations $\delta^*$, such that $\delta^* = \max_{\delta} \ \mathcal{L}(f_s(T(x + \delta)), y); \text{with} \|\delta\|_{\infty} \leq \epsilon$. These techniques vary by their approach ranging from loss function $\mathcal{L}$ design, data augmentation $T(\cdot)$, optimization strategies, or model component targeting. Loss optimization techniques such as SVRE~\cite{xiong2022stochastic}, which takes a stochastic approach, and GRA~\cite{zhu2023boosting}, that updates based on its neighborhood, for optimization, face different challenges. SVRE assumes similar models in the ensemble for variance reduction; however, for most of the crowd models, the similarity ends beyond backbone networks. GRA, on the other hand, overly restricts the perturbation search space. For both approaches, the common limitation is that they cannot handle paradigm shifts in outputs, including heatmaps and coordinates, even though shared backbone vulnerabilities exist.

\noindent Other optimization-based transferable attacks, such as DI$^2$FGSM~\cite{xie2019improving}, Admix~\cite{wang2021admix}, and L2T~\cite{zhu2024learning}, leverage input transformations during optimization. However, DI$^2$FGSM assumes label-preserving linear transformations, which rarely hold in dense point regression tasks such as crowd counting, unlike classification. Admix blends crowd images, introducing artificial patterns and unrealistic distributions. L2T learns dataset-specific transformations that transfer well in classification, but generalize poorly to other tasks. Feature-based approaches like FIA~\cite{wang2021feature}, attack intermediate representations to improve transferability, but their assumption of consistent feature importance fails in crowd counting. This is because, density-map models emphasize spatial distribution, while point-regression models prioritize instance localization, even with shared backbones.

\noindent Generation-based methods synthesize adversarial perturbations (restricted attacks) or full adversarial examples (unrestricted attacks) by removing surrogate-model dependence during inference. Unrestricted methods such as AC-GAN~\cite{song2018constructing}, AT-GAN~\cite{wang2019gan}, and EGM~\cite{xiang2022egm} generate examples from scratch ($\mathcal{L}(f_s(\mathcal{G_\theta}), y)$) but suffer from instability and convergence issues in adversarial training. Diffusion-based approaches, e.g., DiffAttack~\cite{chen2024diffusion}, operate in latent space but often degrade semantic content, producing visually unacceptable results despite strong attacks. Restricted generative methods such as AdvGAN~\cite{xiao2018generating} and GAP~\cite{poursaeed2018generative} instead learn perturbations ($\mathcal{L}(f_s(x + \mathcal{G_\theta}), y)$) under $L_p$ constraints. GE-AdvGAN~\cite{zhu2024ge} further introduces gradient editing and frequency-domain optimization, but still inherits GAN instability. Moreover, frequency optimization alone cannot capture the spatial relationships required for crowd scenes with large scale variations. In summary, existing attacks fail to generalize across paradigms for crowd models, which this work addresses. 
\section{Methodology}\label{sec:meth}
\begin{figure}
    \centering
    \includegraphics[width=\linewidth]{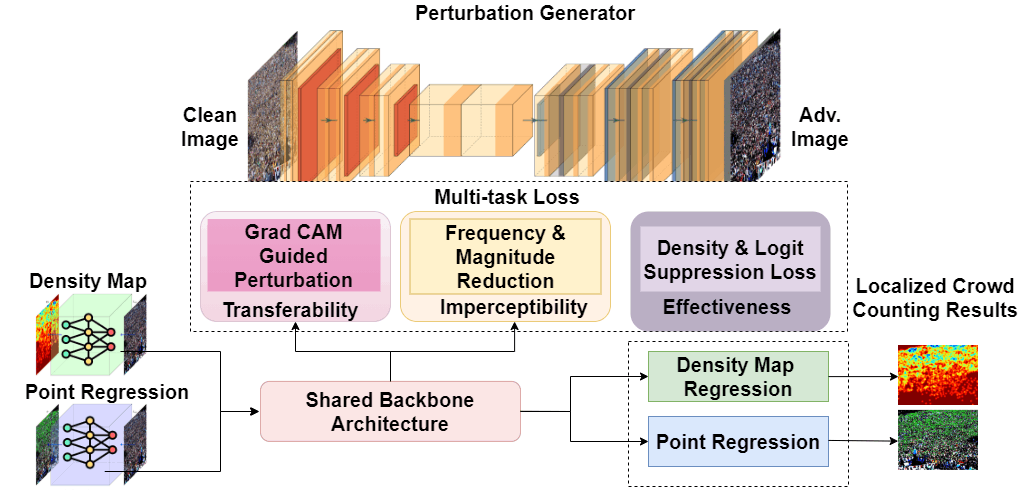}
    \caption{(i) Perturbation loss (GradCAM Perturbation \& Frequency-Magnitude Reduction) in Sec.~\ref{sec:pert}, and (ii) Paradigm-specific losses (Density \& Logit suppression) in Secs.~\ref{sec:logit} and~\ref{sec:den} are proposed for training the perturbation generator $G_\theta$.}
    \label{fig:schematic}
\end{figure}
\subsection{Problem Formulation}
We consider the security of localized crowd counting systems against transferable adversarial attacks. Let $\mathcal{M_P} = \{M_P^1, M_P^2, \dots, M_P^n\}$ denote point regression models that directly output coordinate locations $\mathcal{P} = \{(x_k, y_k)\}_{k=1}^K$ with confidence scores $\mathbf{S}$ from an input image $I \in \mathbb{R}^{H \times W \times 3}$. $\mathcal{M_D} = \{M_D^1, M_D^2, \dots, M_D^m\}$ denote density map models that produce spatial density distributions $\mathcal{D} \in \mathbb{R}^{H \times W}$ from the same image $I$ that later processed into point locations. The complete model set is $\mathcal{M} = \mathcal{M_P} \cup \mathcal{M_D}$.
Our objective is to learn a generator $G_\theta$ that produces perturbations $\delta = G_\theta(I)$ satisfying:
\begin{enumerate}
\item \textbf{Effectiveness}: $I_{\text{adv}} = I + \delta$ causes significant performance degradation across any $\mathcal{M}$.
\item \textbf{Imperceptibility}: $\delta$ is visually subtle with bounded magnitude $\|\delta\|_\infty \leq \epsilon$.
\item \textbf{Cross-paradigm Transferability}: Attacks transfer between $\mathcal{M_D}$ and $\mathcal{M_P}$ leveraging shared backbone vulnerabilities.
\end{enumerate}
\subsection{Generative Adversarial Perturbation Model}
As shown in~\figurename~\ref{fig:schematic}, our framework employs a 3-level U-Net~\cite{ronneberger2015unet} generator $G_\theta$ trained with a multi-task loss consisting of Grad-CAM–guided perturbation, frequency and magnitude regularization, and paradigm-specific density/logit suppression losses. The design exploits shared backbone representations of localized crowd-counting models to improve cross-paradigm transferability and imperceptibility. Prior work shows that backbone networks contribute more to adversarial robustness than detection-specific modules~\cite{li2025importance}. The generator maps an RGB image $I \in \mathbb{R}^{H \times W \times 3}$ to a bounded perturbation $\delta = G_\theta(I)$ with $\|\delta\|_\infty \leq \epsilon$, producing an adversarial example $I_{adv} = \text{clip}(I + \delta)$. Unlike iterative attacks that optimize perturbations per image, the learned mapping $G_\theta : I \rightarrow \delta$ enables a single forward pass at inference time and generalizes across architectures and datasets. Algorithm~\ref{alg:adv_train_generic} summarizes the training procedure. At each iteration, the generator produces perturbations that attack a surrogate model $M_s$ through paradigm-specific loss $\mathcal{L}_{\text{model}}$ while maintaining perceptual constraints via $\mathcal{L}_{\text{pert}}$. The surrogate output $O_{\text{adv}}$ depends on the model type (density map $\mathcal{D}$ or point prediction $\mathcal{P}$). The combined loss $\mathcal{L}_{\text{attack}}$ is then used to update the generator parameters over $E$ epochs.
\begin{algorithm}[t]
\caption{Perturbation Generator Training for Localized Crowd Counting Models}
\label{alg:adv_train_generic}
\SetAlgoLined
\KwIn{Generator $G_\theta$, surrogate model $M_s$, training data $\mathcal{D}$, epochs $E$, bound $\epsilon$, optimizer $\mathcal{O}$}
\KwOut{Trained parameters $\theta$ of $G_\theta$}
\For{$e = 1$ \KwTo $E$}{
    Update learning rate $\eta_e$ using cosine annealing\;
    \ForEach{$(\mathbf{I}, \mathbf{y}) \in \mathcal{D}$}{
        \tcp{Perturbation Generation}
        $\delta \leftarrow G_\theta(\mathbf{I})$\;
        $\delta \leftarrow \text{Clip}(\delta, -\epsilon, \epsilon)$\;
        $\mathbf{I}_{\text{adv}} \leftarrow \text{Clip}(\mathbf{I} + \delta, 0, 1)$\;
        $\mathbf{O}_{\text{adv}} \leftarrow M_s(\mathbf{I}_{\text{adv}})$\;
        \tcp{Loss Computation}
        $\mathcal{L}_{\text{model}} = \alpha\, f_{\text{attack}}(\mathbf{O}_{\text{adv}}, \mathbf{y})$\;
        $\mathcal{L}_{\text{pert}} =
        \beta\, \mathcal{L}_{\text{hinge}}(\delta)
        + \gamma\, \mathcal{L}_{\text{TV}}(\delta)
        + \zeta\, \mathcal{L}_{\text{freq}}(\delta)
        + \kappa\, \mathcal{L}_{\text{cam}}(\mathbf{I}, M_s)$\;
        $\mathcal{L}_{\text{attack}} = \mathcal{L}_{\text{model}} + \mathcal{L}_{\text{pert}}$\;
        \tcp{Parameter update}
        $\nabla_\theta \mathcal{L}_{\text{attack}} \leftarrow 
        \frac{\partial \mathcal{L}_{\text{attack}}}{\partial \theta}$\;
        $\theta \leftarrow \mathcal{O}(\theta, \nabla_\theta \mathcal{L}_{\text{attack}}, \eta_e)$\;
    }
}
\end{algorithm}
\vspace{-2mm}
\subsection{Adversarial Loss Functions}\label{sec:loss}
The intuition of transferable adversarial perturbation optimization is that effective cross-paradigm attacks require both paradigm-specific targeting and universal perceptual constraints. We design a multi-task loss:
\begin{equation}
\mathcal{L}_{\text{attack}} = \underbrace{\mathcal{L}_{\text{model}}}_{\text{paradigm-specific}} + \underbrace{\mathcal{L}_{\text{pert}}}_{\text{cross-paradigm}}
\end{equation}
\begin{equation}
\label{eqn:model_loss}
\mathcal{L}_{\text{model}} =
\begin{cases}
\mathcal{L}_{\text{logit}} & \text{for point-regression models } M_p \\
\mathcal{L}_{\text{dmap}} & \text{for density-map models } M_D
\end{cases}
\end{equation}
where $\mathcal{L}_{\text{model}}$ adapts to surrogate model architecture as shown in Eq.~\ref{eqn:model_loss} while $\mathcal{L}_{\text{pert}}$ ensures perturbations exploit shared latent space across paradigms as shown in \figurename~\ref{fig:schematic}. 
\subsubsection{Logit Suppression Loss}\label{sec:logit}
Point regression models $M_P$ output coordinate locations $\mathcal{P} = \{(x_k, y_k)\}_{k=1}^K$ with logits $\mathbf{L} = \{l_k\}_{k=1}^{K}$ and corresponding confidence scores $\mathbf{S} = \{s_k\}_{k=1}^{K}$. Let $l_i^{(h)}$ denote the human-class logit at position $i$, and let $s_i^{(h)} = \sigma(l_i^{(h)})$ be sigmoid probability. We define the set of high-confidence detection for point coordinates as $\mathcal{P}_{\text{high}} = \{i : s_i^{(h)} > \tau\}$, where $\tau$ is a confidence threshold, either fixed or adaptively decayed over training time as
\begin{equation}
\label{loss:adaptive_th}
\tau(t) = \max\left(\tau_{\min}, \tau_{\max} - \nu \cdot \frac{t}{T_{\max}}\right),
\end{equation}
where $t$ denotes the current epoch, $T_{\max}$ is the total epochs, $\tau_{\min}$ and $\tau_{\max}$ are the lower and upper bounds of $\tau$, and $\nu$ controls the decay rate. The intuition behind adaptive thresholding $\tau$ is to reduce the number of confident logits with training progression. 

\noindent The objective of our \textbf{logit suppression loss} $\mathcal{L}_{\text{logit}}$ is to enforce under-counting by attacking high-confidence detections $\mathcal{P}_{\text{high}}$. Its formulation depends on scene density. For dense scenes ($C_{\text{gt}} > C_{\text{sparse}}$, where $C_{\text{gt}}$ is the ground truth count of that scene and $C_{\text{sparse}}$ is a constant that indicates sparsity threshold containing abundant confident detections, we minimize logits in high-confidence regions using
\begin{equation}
\label{loss:dense_logit}
\mathcal{L}_{\text{dense}} = -\frac{1}{|\mathcal{P}_{\text{high}}|} \sum_{i \in \mathcal{P}_{\text{high}}} \left[ l_i^{(h)} - \log(1 - s_i^{(h)} + \epsilon) \right],
\end{equation}
where we aim to reduce confidence by decreasing logits $l_i^{(h)}$ and increasing $\log(1-s_i^{(h)})$. In contrast, sparse scenes ($C_{\text{gt}} \leq C_{\text{sparse}}$) naturally contain fewer confident detections and require selective penalization around the confidence boundary, given by
\begin{equation}
\label{loss:sparse_logit}
\mathcal{L}_{\text{sparse}} = \frac{1}{|\mathcal{P}_{\text{high}}|} \sum_{i \in \mathcal{P}_{\text{high}}} w_i \cdot l_i^{(h)}, \quad 
w_i = |s_i^{(h)} - \tau|,
\end{equation}
where $w_i = (|s_i^{(h)} - \tau|)$ are spatial weights emphasizing detections near the confidence boundary.
The unified logit suppression loss adapts to scene density by combining equations (\ref{loss:dense_logit}) and (\ref{loss:sparse_logit}):
\begin{equation}
\label{loss:logit}
\mathcal{L}_{\text{logit}} =
\begin{cases}
\begin{aligned}
&\mathcal{L}_{\text{dense}} &\quad \text{Eq. (\ref{loss:dense_logit})}
\end{aligned} & C_{\text{gt}} > C_{\text{sparse}}, \\[3pt]
\begin{aligned}
&\mathcal{L}_{\text{sparse}} &\quad \text{Eq. (\ref{loss:sparse_logit})}
\end{aligned} & C_{\text{gt}} \leq C_{\text{sparse}}.
\end{cases}
\end{equation}

\subsubsection{Density Suppression Loss}\label{sec:den}
\noindent
Density map models $M_D$ generate spatial distributions $\mathcal{D} \in \mathbb{R}^{H \times W}$, from which location predictions $\mathcal{P}$ are derived. We observe that high-density regions with sharp peaks contribute most to count estimates. Thus, we hypothesize that effective density map attacks must target both \emph{absolute intensity} (peak magnitude) and \emph{relative salience} (peak prominence). To evaluate this, we design two complementary formulations of the \textbf{density suppression loss} $\mathcal{L}_{\text{dmap}}$, which is either $\mathcal{L}_{\text{hmap}}$ or $\mathcal{L}_{\text{peak}}$. 

\noindent \textbf{(a) Heatmap Suppression.}  The first formulation $\mathcal{L}_{\text{hmap}}$ jointly attacks absolute peak intensity and near-threshold ambiguity:
\begin{equation}
\label{loss:heat_dmap}
\mathcal{L}_{\text{hmap}} = \underbrace{\frac{1}{|\mathcal{Q^\prime}|} \sum_{(x,y) \in \mathcal{Q^\prime}} \mathcal{D}(x,y)}_{\text{significant peaks}} + \underbrace{\frac{\eta_h}{|\mathcal{Q^*}|} \sum_{(x,y) \in \mathcal{Q^*}} \mathcal{D}(x,y)}_{\text{near-threshold regions}},
\end{equation}
where $\mathcal{Q^\prime}$ denotes significant peak set and $\mathcal{Q^*}$ denotes near threshold set. $\mathcal{Q^\prime}$ and $\mathcal{Q^*}$ are dynamically determined through adaptive thresholding for peak determination in the density map through density value $\mathcal{D}(x,y)$ at each point $(x,y)$. Significant peaks are detected via $3\times3$ max-pooling, yielding local maxima $\chi$. An adaptive threshold $\phi = \phi^\prime \cdot \max(\mathcal{D})$ separates salient peaks from background. $\mathcal{Q^\prime}$ consists of local maxima: $\mathcal{Q^\prime} = \{(x,y) : (\mathcal{D}(x,y) \in \chi) \land (\mathcal{D}(x,y) \geq \phi)\}$, while $\mathcal{Q^*}$ captures regions approaching the threshold: $\mathcal{Q^*} = \{(x,y) : 0.8*\phi \leq \mathcal{D}(x,y) < \phi\}$. 
If $\mathcal{Q^\prime} = \emptyset$, a fallback mechanism selects the top $\mathbf{k}$\% values where $\mathbf{k} = \max(1, \lfloor \phi_0 \cdot |\mathcal{D}| \rfloor)$. The parameter $\eta_h$ controls the relative weight of near-threshold regions.

\noindent \textbf{(b) Peak Suppression.} The second formulation, $\mathcal{L}_{\text{peak}}$, emphasizes relative salience through local prominence:
\begin{equation}
\label{loss:peak_dmap}
\begin{split}
\mathcal{L}_{\text{peak}} = \underbrace{\frac{1}{|\mathcal{Q^\prime}|} \sum_{(x,y) \in \mathcal{Q^\prime}} \mathcal{D}(x,y)}_{\text{peak magnitude}} + \\
 \underbrace{\frac{\eta_p}{|\mathcal{Q^\prime}|} \sum_{(x,y) \in \mathcal{Q^\prime}} \left[\mathcal{D}(x,y) - \mu_{\text{loc}}(x,y)\right]}_{\text{peak prominence}}
\end{split}
\end{equation}
where $\mu_{\text{loc}}(x,y)$ denotes local neighborhood maxima computed via $3\times3$ max-pooling.

\noindent The selection of $\mathcal{L}_{\mathrm{hmap}}$ and $\mathcal{L}_{\mathrm{peak}}$ is based on the density of people located in an area. $\mathcal{L}_{\mathrm{peak}}$ efficiently targets isolated high-density crowd regions, while $\mathcal{L}_{\mathrm{hmap}}$ targets uniformly distributed crowd. We choose $\mathcal{L}_{\mathrm{peak}}$ over $\mathcal{L}_{\mathrm{hmap}}$ by isolation ratio $>0.7$, which is computed as the fraction of peaks without neighboring peaks within a $5\times5$ window.

\subsubsection{Perturbation Loss} \label{sec:pert}
The perturbation loss $\mathcal{L}_{\text{pert}}$ ensures adversarial examples are both effective and imperceptible while enhancing cross-paradigm transferability. We design a multi-constraint approach that balances attack efficacy and perceptibility. Different localized crowd counting models apply similar architectures that share common backbones or feature extractor networks. We assume the common backbones learns a similar latent space with similar inductive biases~\cite{conwell2024large}. Based on this assumption, we design multi-task optimization objectives that exploit universal vulnerabilities rather than model-specific characteristics. Specifically, our multi-task objective includes a frequency-domain constraint, a magnitude constraint, gradient-weighted activation guidance, and a spatial smoothness regularization term.

\noindent\textbf{Frequency Domain Constraint:}
We observe that crowd scenes exhibit strong low-frequency dominance due to smooth human shapes and organic distributions. High-frequency perturbations create artificial patterns that are both easily detectable and poorly transferable across architectures with different filtering and down-sampling operations. Therefore, we design a frequency loss $\mathcal{L}_{\text{freq}}$ that suppresses high-frequency components to align perturbations with natural crowd statistics is computed on the frequency domain using Fast Fourier transform (FFT) on $\delta$.
\begin{equation}
\mathcal{L}_{\text{freq}} = \frac{1}{|\Omega|} \sum_{\omega \in \Omega} |\mathcal{F}(\delta)(\omega)|,
\label{eqn:freq}
\end{equation}
where $\Omega$ excludes the average value of the signal $\delta$, ($\omega \neq 0$) and $\mathcal{F}$ denotes FFT. This ensures attacks focus on semantically meaningful low-frequency features that generalize across paradigms.

\noindent\textbf{GradCAM-Guided Perturbation:}
GradCAM~\cite{selvaraju2017grad} is an explainable AI technique used to produce visual explanations for decisions made by neural networks. We propose the GradCAM-guided loss $\mathcal{L}_{\text{cam}}$, which leverages shared backbone activations to focus perturbations on semantically important regions that generalize across paradigms. Instead of perturbing the whole image equally, we propose to minimize the perturbation that falls outside the parts of the image regarded as important by GradCAM.
\begin{equation}
\mathcal{L}_{\text{cam}} = \frac{1}{HW} \left\| \, |\delta| - \delta(\rho) \, \right\|_1,
\label{eqn:cam}
\end{equation}
where $\rho$ is the CAM attention map. By minimizing perturbations outside attention regions, we enhance transferability through shared feature vulnerabilities.

\noindent\textbf{Magnitude Constraint:} 
Hinge loss $\mathcal{L}_{\text{hinge}}$ bounds overall perturbation energy to maintain visual fidelity by taking $L_2$ norm of perturbation magnitude $\delta$, where $\delta \in \mathbb{R}^{H\times W}$.
\begin{equation}
\begin{split}
\mathcal{L}_{\text{hinge}} = \mathbb{E}[\delta^2] = \frac{1}{HW} \|\delta\|_2^2 = \frac{1}{HW} \sum_{i=1}^H \sum_{j=1}^W \delta_{i,j}^2
\end{split}
\label{eqn:hinge}
\end{equation}
\noindent\textbf{Spatial Smoothness Regularization:}
The total variation loss $\mathcal{L}_{\text{tv}}$ is employed to normalize the total variations of perturbation magnitude $\delta$. 
\begin{equation}
\mathcal{L}_{\text{tv}} = \frac{1}{HW} \left[ \sum_{i,j} |\delta_{i,j+1} - \delta_{i,j}| + \sum_{i,j} |\delta_{i+1,j} - \delta_{i,j}| \right]
\label{eqn:tv}
\end{equation}

The final multi-task loss ensures subtle but meaningful perturbations as well as transferability, which combines the above cross-paradigm constraints in Eq.~\ref{eqn:freq}, Eq.~\ref{eqn:cam}, Eq.~\ref{eqn:hinge}, and Eq.~\ref{eqn:tv}, as shown in Eq.~\ref{eqn:me}. 
\begin{equation}
\mathcal{L}_{pert}= \beta \cdot \mathcal{L}_{\text{hinge}} + \gamma \cdot \mathcal{L}_{\text{tv}} + \zeta \cdot \mathcal{L}_{\text{freq}} + \kappa \cdot \mathcal{L}_{\text{cam}}
\label{eqn:me}
\end{equation}

\section{Experimental Settings}\label{sec:exp}
\subsection{Dataset}
For evaluation, we use two widely adopted crowd-counting datasets. (i) Shanghai Tech Part A (SHHA)~\cite{zhang2016single} consists of 482 images with, on average, 501 people per image. (ii) UCF-QNRF~\cite{idrees2018composition} consists of 1,535 high-resolution crowd images with, on average, 815 people per image. They often contain instances of extremely dense crowds; for example, UCF-QNRF contains 12,835 people in a single image, providing suitable examples to test our hypothesis. For each dataset, the training set is used to train $G_\theta$, and the test set is used for adversarial example evaluation. Input images are resized to $512 \times 512$ resolution during the training stage.

\subsection{Baselines}
\noindent We compare our method with representative adversarial attacks from both optimization- and generation-based approaches. Optimization-based baselines include data-augmentation methods DI$^2$FGSM~\cite{xie2019improving}, Admix~\cite{wang2021admix}, and L2T~\cite{zhu2024learning}; loss-based methods SVRE~\cite{xiong2022stochastic} and GRA~\cite{zhu2023boosting}; and the feature-based method FIA~\cite{wang2021feature}. For generative attacks, we evaluate against DiffAttack~\cite{chen2024diffusion} and GE-AdvGAN~\cite{zhu2024ge}. We additionally compare with PAP~\cite{liu2022harnessing}, a recent attack designed for crowd counting models.
\subsection{Evaluation Metrics}
We evaluate three objectives: \textit{attack strength}, \textit{transferability}, and \textit{imperceptibility}. 
Attack strength is measured by Mean Absolute Error (MAE) and Miss Rate (MR); transferability by the MAE Ratio (TR) between target and surrogate models; and imperceptibility by Structural Similarity Index (SSIM) and Peak Signal-to-Noise Ratio (PSNR).

\noindent\textbf{MAE} measures the average counting error:
\begin{equation}
\text{MAE} = \frac{1}{N} \sum_{i=1}^{N} |C_{\text{clean}}^{(i)} - C_{\text{adv}}^{(i)}|.
\end{equation}

\noindent\textbf{MR} denotes the fraction of missed individuals:
\begin{equation}
\text{MR} = \frac{1}{N} \sum_{i=1}^{N} 
\frac{C_{\text{clean}}^{(i)} - C_{\text{adv}}^{(i)}}{C_{\text{clean}}^{(i)}}.
\end{equation}

\noindent\textbf{TR} evaluates transferability:
\begin{equation}
\text{TR} = \frac{\text{MAE}_{\text{target}}}{\text{MAE}_{\text{surrogate}}}.
\end{equation}

\noindent\textbf{SSIM} and \textbf{PSNR} measure perceptual similarity between the clean image $I$ and adversarial image $I_{\text{adv}}$:
\begin{equation}
\text{SSIM}(I,I_{\text{adv}})=
\frac{(2\mu_I\mu_{I_{\text{adv}}}+c_1)(2\sigma_{II_{\text{adv}}}+c_2)}
{(\mu_I^2+\mu_{I_{\text{adv}}}^2+c_1)(\sigma_I^2+\sigma_{I_{\text{adv}}}^2+c_2)}.
\end{equation}
\begin{equation}
\text{PSNR}(I,I_{\text{adv}})=
10\log_{10}\left(\frac{\text{MAX}_I^2}{\text{MSE}(I,I_{\text{adv}})}\right).
\end{equation}
where $\text{MAX}_I=255$ for 8-bit images.

\subsection{Surrogate and Target Models}
\noindent We evaluate our attack across eight crowd-counting models spanning point-regression and density-map paradigms. The point-regression group includes P2PNet~\cite{song2021rethinking}, PET~\cite{liu2023point}, and APGCC~\cite{chen2024improving}, while the density-map group includes SASNet~\cite{song2021choose}, FIDTM~\cite{liang2022focal}, HMoDE~\cite{du2023redesigning}, and Gramformer~\cite{lin2024gramformer}. Among these, PET, APGCC, and Gramformer are transformer-based, while the others are convolutional. Most models use VGG-16, VGG-19, or ResNet-50 backbones, except FIDTM, which employs HRNet.

\subsection{Hyperparameter in Multi-task Loss}
All the experiments are conducted on $512 \times 512$ for the sake of avoiding memory run-out errors, which is standard practice. Attack budget is $\epsilon = 8/255$ for all the methods.The multi-task loss in Eq.~\ref{eqn:me} employs weights that are calibrated through grid search on validation set to balance attack strength and stealth. The hinge loss weight $\beta = 0.01$, total variation weight $\gamma = 0.05$, frequency domain constraint $\zeta = 0.01$, and GradCAM attention guidance $\kappa = 0.5$. For logit suppression loss, the sparse scene hyperparameter $C_{\text{sparse}}$ is set to 100. $\alpha$ is set to between $1.0$ to $C_{\text{gt}}/100$ for both approaches. The adaptive threshold parameters follow $\tau_{\min} = 0.3$, $\tau_{\max} = 0.5$ which are typical confidence ranges in crowd models. The decay rate $\nu = 0.2$ avoids abrupt threshold changes during training.
\section{Evaluation}
\begin{figure}
    \centering
    \includegraphics[width=\linewidth]{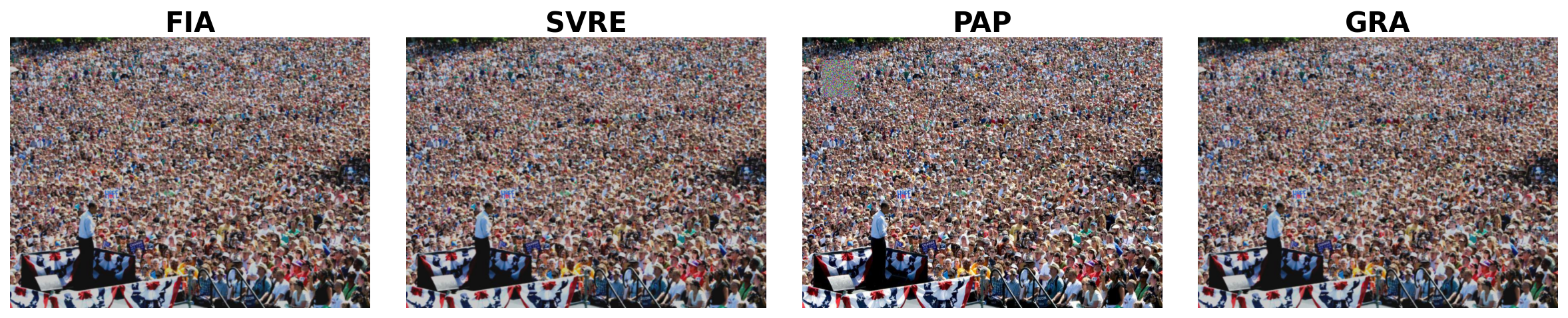}
    \includegraphics[width=\linewidth]{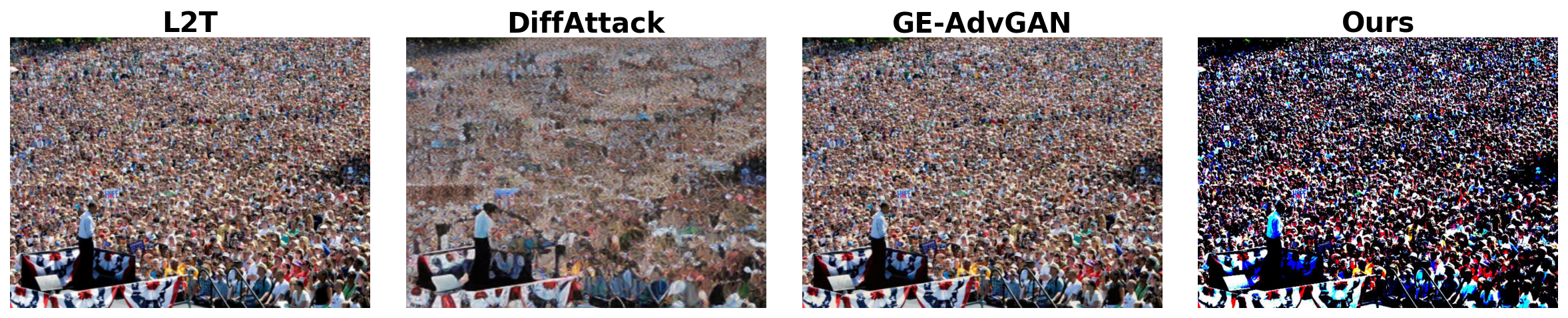}
    \caption{Sample adversarial image comparison with our method with respect to state-of-the-art models}
    \label{fig:adv_collage}
    \vspace{-2mm}
\end{figure}

\subsection{Attack Transferability}
We first evaluate the cross-paradigm transferability of our designed attack across seven localized crowd counting models on two different datasets, as shown in Tables~\ref{tab:shha_transfer} and~\ref{tab:ucf_transfer}. From the results, we show three main critical patterns that emerge from the transfer ratios (TR) across both datasets.

\begin{table*}[!t]
\centering
\caption{Transferability of our designed adversarial examples across crowd counting models \textbf{(MAE/TR) on the SHHA} dataset., where the highest and lowest values are indicated by \textbf{bold} and \underline{underline}.}
\label{tab:shha_transfer}
\small
\begin{tabular}{l|ccccccc}
\hline
\multirow{2}{*}{\textbf{Source Model}} & \multicolumn{7}{c}{\textbf{Target Model}} \\
\cline{2-8}
 & \textbf{SASNet} & \textbf{P2PNet} & \textbf{FIDTM} & \textbf{PET} & \textbf{HMoDE} & \textbf{APGCC} & \textbf{Gramformer} \\
\hline
\textbf{SASNet}~\cite{song2021choose} & 300.82/1.0 & 420.17/1.40 & 338.67/1.13 & 364.59/1.21 & 249.61/0.83 & 397.96/1.32 & 257.54/0.86\\
\textbf{P2PNet}~\cite{song2021rethinking} & 281.00/0.89 & 316.58/1.0 & 268.90/0.85 & 255.61/0.81 & 189.79/0.60 & 224.67/0.71 & 221.08/0.70\\
\textbf{FIDTM}~\cite{liang2022focal} & 313.45/1.21 & 426.89/1.64 & 353.08/1.0 & 373.70/1.44 & 250.52/0.96 & 407.76/1.57 & 277.25/1.07\\
\textbf{PET}~\cite{liu2023point} & 315.00/1.03 & 430.91/1.41 & 350.84/1.15 & 305.43/1.0 & 252.33/0.83 & 406.19/1.33 & 269.92/0.88\\
\textbf{HMoDE}~\cite{du2023redesigning} & 301.36/1.21 & 420.71/$\textbf{1.69}$ & 338.82/1.36 & 366.16/1.47 & 249.19/1.0 & 397.48/1.60 & 257.62/1.03\\
\textbf{APGCC}~\cite{chen2024improving} & 268.03/0.86 & 297.82/0.95 & 257.32/0.82 & 231.59/0.74 & 171.53/\underline{0.55} & 313.45/1.0 & 203.11/0.65\\
\textbf{Gramformer}~\cite{lin2024gramformer} & 297.73/1.17 & 419.22/1.64 & 338.8/1.33 & 364.24/1.43 & 249.11/0.98 & 398.49/1.56 & 254.89/1.0\\
\hline
\textbf{Clean} & 36.98 & 46.12 & 28.10 & 70.21 & 74.80 & 56.55 & 29.91\\
\hline
\end{tabular}
\vspace{-2mm}
\end{table*}

\begin{table*}[!t]
\centering
\caption{Transferability of our designed adversarial examples across crowd counting models \textbf{(MAE/TR) on UCF-QNRF} dataset, where the highest and lowest values are indicated by \textbf{bold} and \underline{underline}.}
\label{tab:ucf_transfer}
\small
\begin{tabular}{l|ccccccc}
\hline
\multirow{2}{*}{\textbf{Source Model}} & \multicolumn{7}{c}{\textbf{Target Model}} \\
\cline{2-8}
 & \textbf{SASNet} & \textbf{P2PNet} & \textbf{FIDTM} & \textbf{PET} & \textbf{HMoDE} & \textbf{APGCC} & \textbf{Gramformer} \\
\hline
\textbf{SASNet}~\cite{song2021choose} & 619.74/1.0 & 819.59/1.32 & 741.62/1.20 & 843.36/1.36 & 526.73/0.85 & 808.79/1.31 & 613.61/0.99\\
\textbf{P2PNet}~\cite{song2021rethinking} & 615.04/0.93 & 663.95/1.0 & 632.00/0.95 & 671.59/1.01 & 437.98/\underline{0.66} & 686.58/1.03 & 527.64/0.79\\
\textbf{FIDTM}~\cite{liang2022focal} & 639.64/0.83 & 833.09/1.09 & 767.54/1.0 & 846.11/1.10 & 530.23/0.69 & 819.93/1.07 & 647.00/0.84\\
\textbf{PET}~\cite{liu2023point} & 630.50/0.83 & 831.23/1.09 & 747.29/0.98 & 763.30/1.0 & 530.11/0.69 & 813.33/1.07 & 621.62/0.81\\
\textbf{HMoDE}~\cite{du2023redesigning} & 619.96/1.18 & 819.81/1.55 & 741.44/1.41 & 843.42/$\textbf{1.60}$ & 527.32/1.0 & 808.13/1.53 & 614.00/1.16\\
\textbf{APGCC}~\cite{chen2024improving} & 634.10/0.79 & 828.21/1.03 & 744.52/0.92 & 762.85/0.95 & 538.86/0.67 & 805.30/1.0 & 616.80/0.77\\
\textbf{Gramformer}~\cite{lin2024gramformer} & 620.00/1.0 & 820.31/1.34 & 741.87/1.21 & 843.33/1.37 & 528.07/0.86 & 808.04/1.32 & 613.91/1.0\\
\hline
\textbf{Clean} & 326.94 & 324.73 & 279.08 & 393.19 & 339.02 & 288.12 & 290.44\\
\hline
\end{tabular}
\end{table*}

\noindent First, we identify cases of \emph{super-transferability} where attacks perform \emph{better} on target models than their source (TR $> 1.0$). Notably, \textbf{HMoDE} attacks achieve $69\%$ higher effectiveness on \textbf{P2PNet} than on itself (i.e., TR = $1.69$), indicating that black-box attackers can potentially outperform white-box attackers against certain model combinations due to complex head architecture (e.g., multi-scale gating) amplifying the shared backbone vulnerabilities. 

\noindent Second, transferability patterns are consistent across datasets. For instance, \textbf{HMoDE} emerges as the strongest surrogate model (SHHA: TR up to $1.69$, UCF-QNRF: TR up to $1.60$) and \textbf{P2PNet} as the most vulnerable target across both benchmarks. Surprisingly, CNN-based HMoDE achieves strong transfer to transformer-based PET (TR=1.60 on UCF-QNRF, TR=1.47 on SHHA), which suggests that a shared backbone but different architecture paradigm learn a similar latent space based on our proposed attack framework.

\begin{figure}
    \centering
    \includegraphics[width=0.22\linewidth]{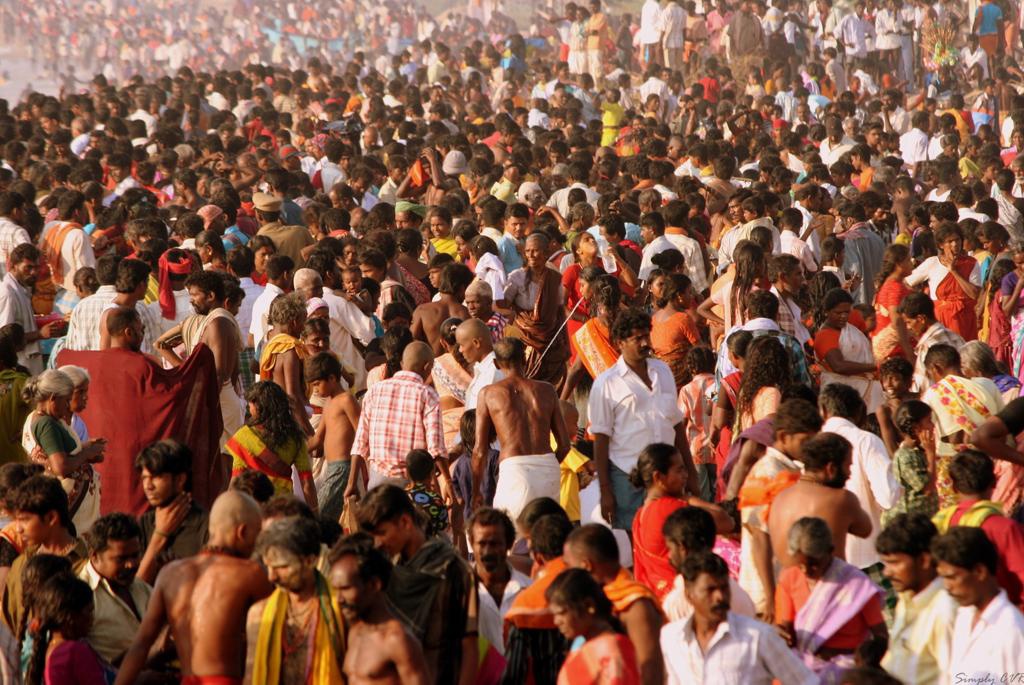}
    \hspace{0.02\linewidth}
    \includegraphics[width=0.22\linewidth]{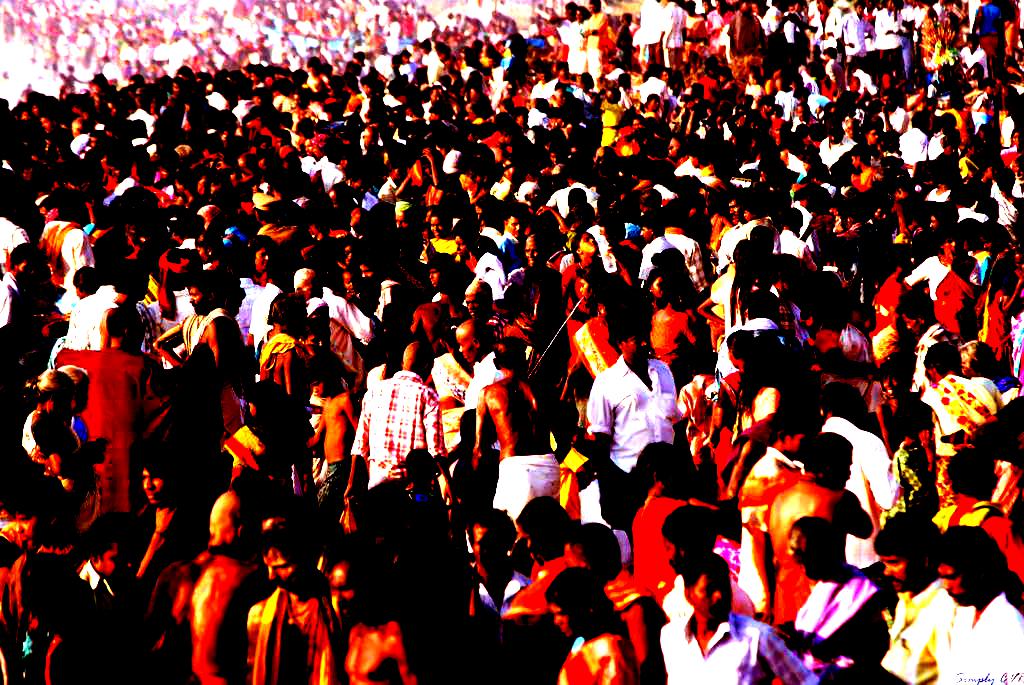}
    \hspace{0.02\linewidth}
    \includegraphics[width=0.22\linewidth]{reb/clean_IMG_297.jpg}
    \hspace{0.02\linewidth}
    \includegraphics[width=0.22\linewidth]{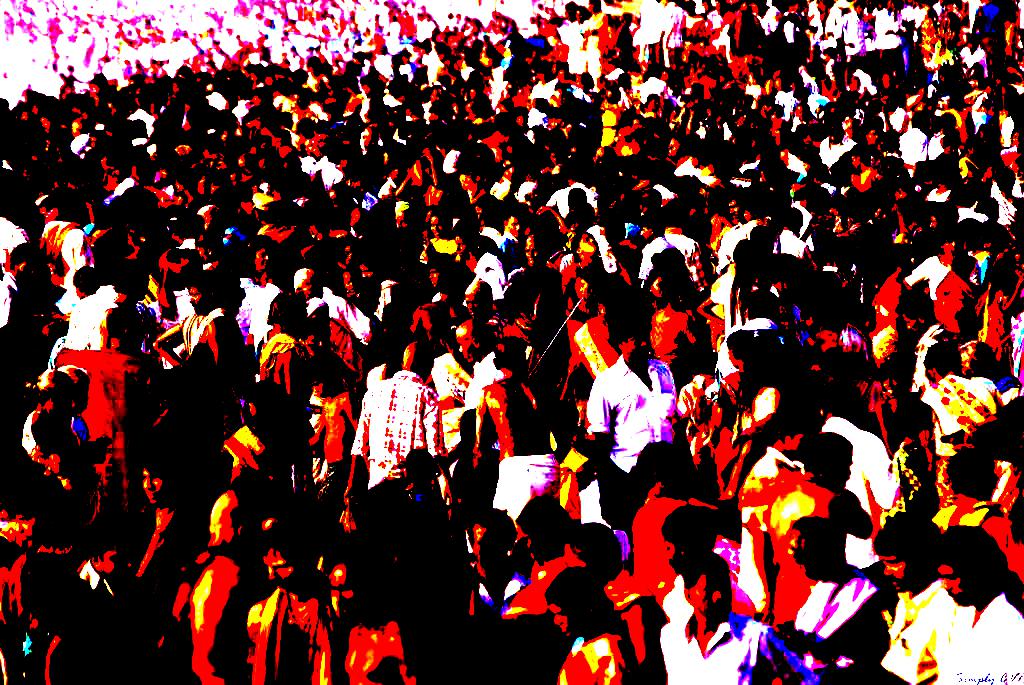}

    \includegraphics[width=0.22\linewidth]{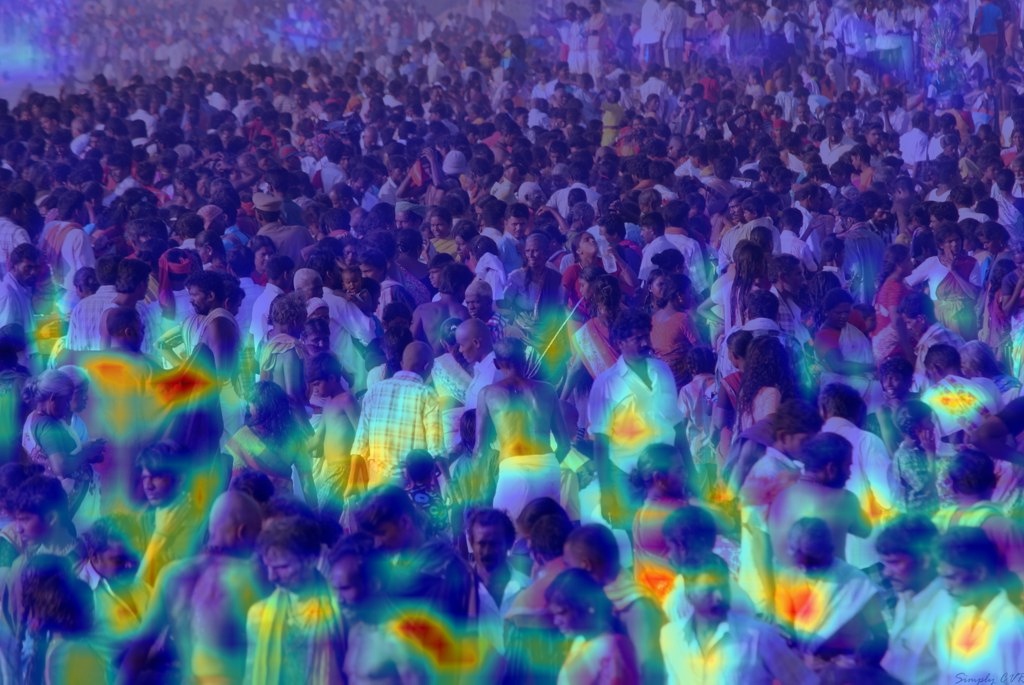}
    \hspace{0.02\linewidth}
    \includegraphics[width=0.22\linewidth]{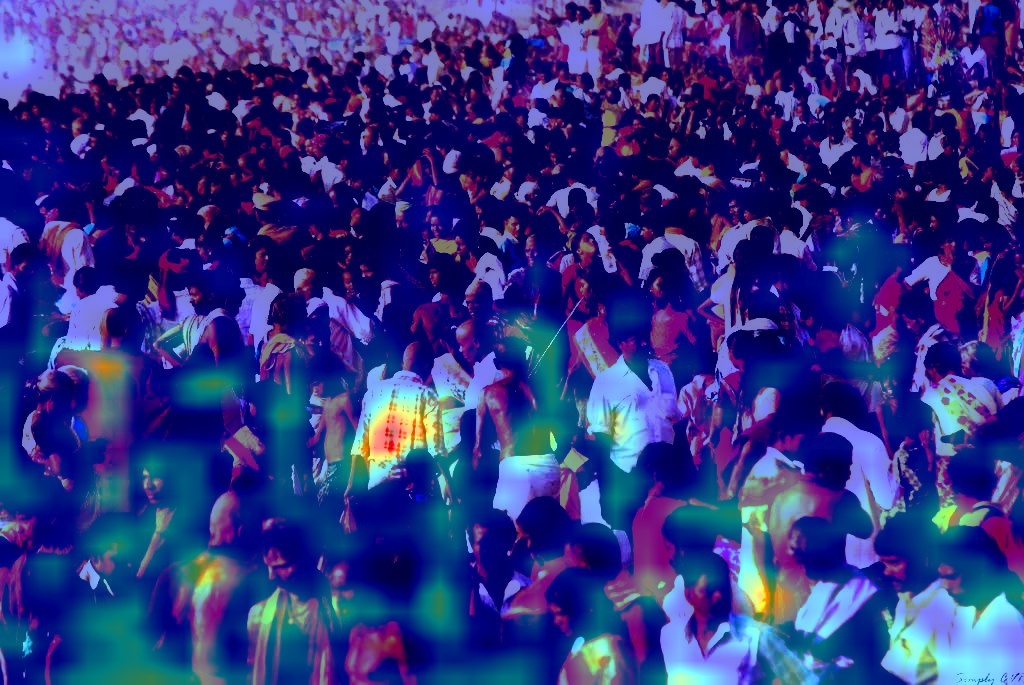}
    \hspace{0.02\linewidth}
    \includegraphics[width=0.22\linewidth]{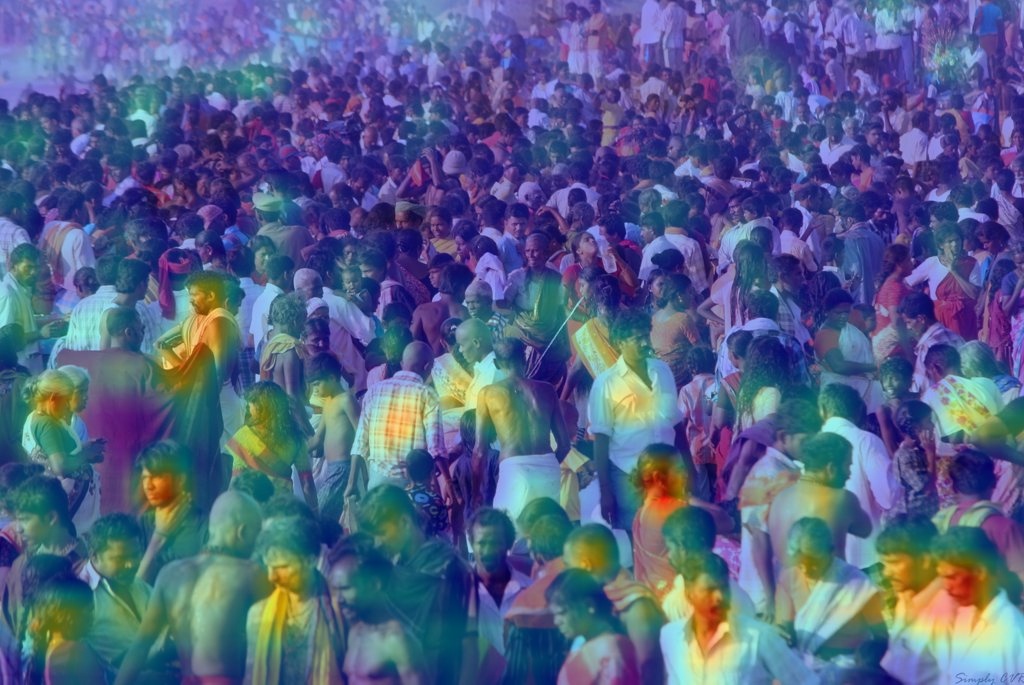}
    \hspace{0.02\linewidth}
    \includegraphics[width=0.22\linewidth]{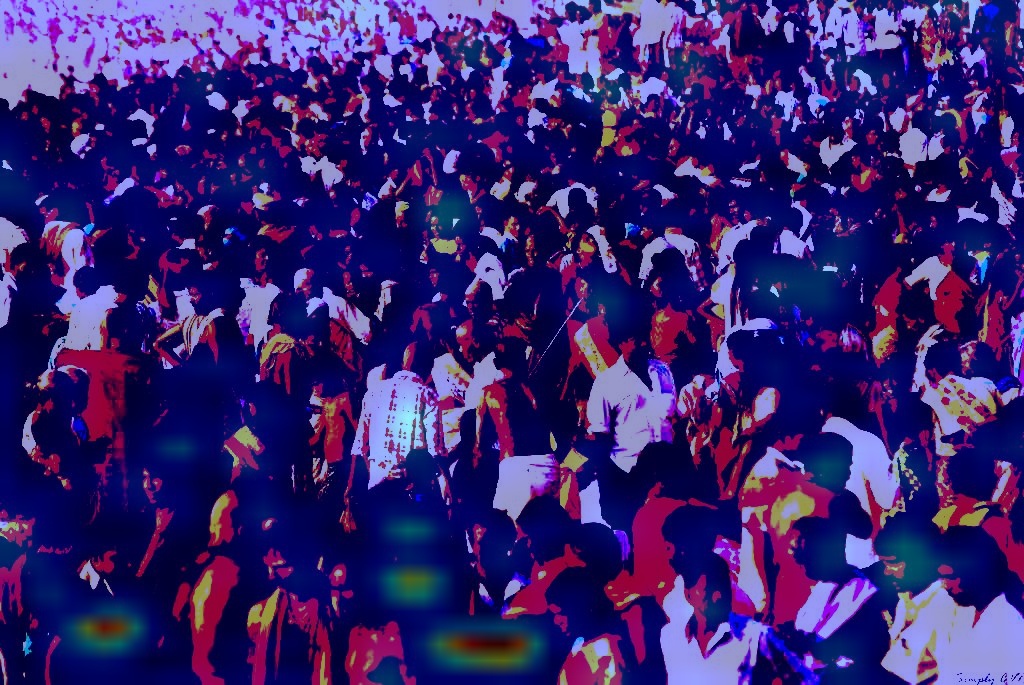}
    
    \parbox{0.22\linewidth}{\centering \scriptsize (a) Clean (P2PNet)}
    \hspace{0.02\linewidth}
    \parbox{0.22\linewidth}{\centering \scriptsize (b) Adversarial (P2PNet)}
    \hspace{0.02\linewidth}
    \parbox{0.22\linewidth}{\centering \scriptsize (c) Clean (SASNet)}
    \hspace{0.02\linewidth}
    \parbox{0.22\linewidth}{\centering \scriptsize (d) Adversarial (SASNet)}
    \caption{GradCAM response of clean and adversarial images} 
    \label{fig:cam}
    \vspace{-10pt}
\end{figure}

\noindent Third, our method achieves strong cross-paradigm transferability between density-map and point-regression architectures. On SHHA, attacks from density-based \textbf{HMoDE} achieve TR=$1.69$ against point-regression \textbf{P2PNet}, while point-regression \textbf{PET} transfers effectively to density-based \textbf{FIDTM} (TR=$1.15$). This cross-paradigm success is particularly significant as it demonstrates that adversarial vulnerabilities transcend fundamental architectural differences. The universal vulnerability is evident with all models showing susceptibility to cross-paradigm attacks (TR $> 0.6$ in most cases). Even the weakest transfer case (\textbf{APGCC} to \textbf{HMoDE}, TR=$0.55$) still represents significant security compromise, increasing MAE from $74.80$ to $171.53$. The Grad-CAM visualizations (Fig.~(a-b) for point-based, Fig.~(c-d) for density-based) in Figure~\ref{fig:cam} show thinned-out activation of semantically important crowd regions in adversarial images.

\noindent These results reveal security gaps in crowd counting where attacks transfer across architectural boundaries due to shared inductive biases~\cite{baxter2000model} (e.g., VGG-16 and ResNet-50), lowering the barrier to real-world exploitation.

\subsection{Baseline Comparison}
We evaluate the performance of our designed attack for localized counting models with nine state-of-the-art transferable attack methods on the SHHA dataset, as shown in Table~\ref{tab:complete_results}. On one extreme, \textbf{PAP} and \textbf{GE-AdvGAN} maintain high visual quality (PSNR $>22$ dB, SSIM $>0.87$) but achieve low attack success (MAE $<120$), while \textbf{DiffAttack} delivers very strong attacks (MAE $414.66$) at the cost of low quality (PSNR $11.52$ dB). Our method strikes the optimal balance, achieving $259.76$ MAE—$3.7\times$ more effective than high-quality methods—while maintaining $19.25$ dB PSNR, representing a $67\%$ quality improvement over DiffAttack's extreme approach.
\begin{table*}[!t]
\centering
\caption{Evaluation of attacks using FIDTM model across density regimes (SHHA dataset), where $\uparrow$ indicates the higher the better.}
\label{tab:complete_results}
\resizebox{\textwidth}{!}{%
\begin{tabular}{l|cccc|cccc|cccc|cccc}
\hline
 & \multicolumn{4}{c|}{\textbf{Overall}} & \multicolumn{4}{c|}{\textbf{Sparse (\textless100)}} & \multicolumn{4}{c|}{\textbf{Moderate (100-1000)}} & \multicolumn{4}{c}{\textbf{Dense (\textgreater1000)}} \\
\cline{2-17}
\textbf{Method} & \textbf{MAE}$\uparrow $ & \textbf{MR}$\uparrow $ & \textbf{PSNR}$\uparrow $ & \textbf{SSIM}$\uparrow $ & \textbf{MAE}$\uparrow $ & \textbf{MR}$\uparrow $ & \textbf{PSNR}$\uparrow $ & \textbf{SSIM}$\uparrow $ & \textbf{MAE}$\uparrow $ & \textbf{MR}$\uparrow $ & \textbf{PSNR}$\uparrow $ & \textbf{SSIM}$\uparrow $ & \textbf{MAE}$\uparrow $ & \textbf{MR}$\uparrow $ & \textbf{PSNR}$\uparrow $ & \textbf{SSIM}$\uparrow $ \\
\hline
Clean & 28.1 & 4.28 & N/A & N/A & 2.83 & 5.06 & N/A & N/A & 16.69 & 3.75 & N/A & N/A & 128.0 & 7.82 & N/A & N/A \\
\hline
DI-$^2$FGSM \cite{xie2019improving} & 77.14 & 11.44 & 20.59 & 0.836 & 4.33 & 6.09 & 23.65 & 0.924 & 56.44 & 11.30 & 20.78 & 0.843 & 273.0 & 15.47 & 17.43 & 0.736 \\
Admix \cite{wang2021admix} & 69.21 & 9.73 & 20.80 & 0.841 & 2.83 & 4.60 & 24.14 & 0.930 & 48.61 & 9.44 & 21.00 & 0.848 & 260.8 & 14.71 & 17.48 & 0.738 \\
FIA \cite{wang2021feature} & 68.96 & 10.27 & 20.59 & 0.837 & 3.67 & 5.23 & 23.65 & 0.924 & 50.88 & 10.18 & 20.78 & 0.844 & 241.0 & 13.73 & 17.43 & 0.736 \\
SVRE \cite{xiong2022stochastic} & 77.83 & 11.76 & 20.58 & 0.836 & 4.83 & 6.72 & 23.64 & 0.924 & 58.35 & 11.68 & 20.78 & 0.843 & 264.6 & 15.11 & 17.43 & 0.736 \\
GRA \cite{zhu2023boosting} & 69.87 & 10.35 & 20.60 & 0.837 & 5.83 & 8.90 & 23.66 & 0.924 & 50.46 & 9.93 & 20.80 & 0.844 & 251.3 & 14.32 & 17.44 & 0.737 \\
L2T \cite{zhu2024learning} & 70.17 & 9.60 & 20.79 & 0.842 & 2.83 & 4.56 & 24.07 & 0.931 & 48.76 & 9.26 & 20.99 & 0.849 & 268.5 & 14.90 & 17.49 & 0.740 \\
GE-AdvGAN \cite{zhu2024ge} & 110.78 & 15.66 & 23.20 & 0.875 & 5.17 & 7.65 & 24.20 & 0.922 & 64.33 & 14.19 & 23.36 & 0.878 & 518.8 & 31.11 & 21.48 & 0.827 \\
DiffAttack \cite{chen2024diffusion} & 414.66 & 67.13 & 11.52 & 0.483 & 14.33 & 25.07 & 12.88 & 0.577 & 287.66 & 67.36 & 11.58 & 0.496 & 1591.3 & 88.37 & 10.36 & 0.332 \\
PAP \cite{liu2022harnessing} & 32.72 & 5.81 & 22.46 & 0.976 & 4.67 & 7.42 & 19.13 & 0.954 & 22.37 & 5.40 & 22.58 & 0.977 & 126.1 & 8.10 & 23.32 & 0.982 \\
\hline
\textbf{Ours using SASNet} & 259.76 & 50.01 & 19.25 & 0.820 & 14.33 & 21.93 & 18.99 & 0.819 & 177.60 & 45.36 & 19.33 & 0.822 & 1013.6 & 58.42 & 18.81 & 0.806 \\
\textbf{Ours using P2PNet} & 271.71	& 47.72 & 19.18 & 0.817 & 12.83 & 18.62 & 18.91 & 0.8167 & 188.58 & 48.19 & 19.25 & 0.8189 & 1040.18 & 60.04 & 18.73 & 0.8026\\
\hline
\end{tabular}%
}
\end{table*}
\noindent In dense scenes (defined as more than 1,000 people), our attack increases MAE from 128.0 to 1,013.6 while maintaining 18.81 dB PSNR, showing counting failure induction without visual collapse. This contrasts with DiffAttack's complete quality collapse (PSNR $10.36$ dB) for similar effectiveness. SASNet and P2PNet represent density-map and logit-suppression variants that deliver consistent performance, with P2PNet achieving slightly higher MAE ($271.71$) while maintaining comparable quality, validating our method's robustness across different surrogate architectures. Our method demonstrates density-aware effectiveness, achieving $58.42\%$ miss rate in dense scenes, hiding the majority of a crowded population, while previous methods plateau at $15$-$31\%$ miss rates. 

\noindent To summarize, compared to existing state-of-the-art methods, our work achieves a better balance between attack efficacy, transferability, and imperceptibility: we deliver $9\times$ MAE degradation while maintaining deployable $19$ dB visual quality, redefining the attack frontier for real-world scenarios where detection avoidance is paramount.
\subsection{Ablation Studies}
\begin{table}[!t]
\centering
\caption{Impact of combinations of loss functions on MR and PSNR for SHHA dataset.}
\label{tab:ablation}
\small
\begin{tabular}{@{}p{5.5cm}|c|c@{}}
\toprule
Loss Equations & MR(\%) & PSNR \\
\midrule
$\alpha\mathcal{L}_\text{hmap} + \beta\mathcal{L}_\text{hinge}$ & 45.35 & 17.67 \\
$\alpha\mathcal{L}_\text{hmap} + \beta\mathcal{L}_\text{hinge}+\kappa\mathcal{L}_\text{cam}$ & 59.47 &  17.67\\
$\alpha\mathcal{L}_\text{hmap} + \beta\mathcal{L}_\text{hinge}+\zeta\mathcal{L}_\text{freq}$ & 60.46 & \textbf{17.75}  \\
$\alpha\mathcal{L}_\text{hmap} + \beta\mathcal{L}_\text{hinge}+\gamma\mathcal{L}_\text{tv}+\zeta\mathcal{L}_\text{freq}$ & 60.26 & 17.65 \\
$\alpha\mathcal{L}_\text{hmap} + \beta\mathcal{L}_\text{hinge}+\gamma\mathcal{L}_\text{tv}+\zeta\mathcal{L}_\text{freq}+ \kappa\mathcal{L}_\text{cam}$ & \textbf{60.89} & 17.47 \\
\midrule
$\alpha\mathcal{L}_\text{peak}+\beta\mathcal{L}_\text{hinge}$ & 44.93 & 17.71 \\
$\alpha\mathcal{L}_\text{peak}+\beta\mathcal{L}_\text{hinge}+\kappa\mathcal{L}_\text{cam}$ & 60.36 & 17.50 \\
$\alpha\mathcal{L}_\text{peak}+\beta\mathcal{L}_\text{hinge}+\zeta\mathcal{L}_\text{freq}$ & 60.46 & 17.63 \\
$\alpha\mathcal{L}_\text{peak}+\beta\mathcal{L}_\text{hinge}+\gamma\mathcal{L}_\text{tv}+\zeta\mathcal{L}_\text{freq}$ & 58.11 & \textbf{17.80} \\
$\alpha\mathcal{L}_\text{peak}+\beta\mathcal{L}_\text{hinge}+\gamma\mathcal{L}_\text{tv}+\zeta\mathcal{L}_\text{freq}+ \kappa\mathcal{L}_\text{cam}$ & \textbf{60.90} & 17.46 \\
\midrule
$\alpha\mathcal{L}_\text{logit}+ \beta\mathcal{L}_\text{hinge}$ & 45.15 & 19.09 \\
$\alpha\mathcal{L}_\text{logit}+ \beta\mathcal{L}_\text{hinge}+ \kappa\mathcal{L}_\text{cam}$ & \textbf{45.61} & \textbf{19.10} \\
$\alpha\mathcal{L}_\text{logit}+\beta\mathcal{L}_\text{hinge}+\zeta\mathcal{L}_\text{freq}$ & 43.74 & 18.98 \\
$\alpha\mathcal{L}_\text{logit}+\beta\mathcal{L}_\text{hinge}+\gamma\mathcal{L}_\text{tv}+\zeta\mathcal{L}_\text{freq}$ & 41.35 & 19.04 \\
$\alpha\mathcal{L}_\text{logit}+ \beta\mathcal{L}_\text{hinge}+\gamma\mathcal{L}_\text{tv}+\zeta\mathcal{L}_\text{freq}+ \kappa\mathcal{L}_\text{cam}$ & 45.05 & 19.07 \\
\bottomrule
\end{tabular}
\vspace{-2mm}
\end{table}
We evaluate our loss function design choices of various combinations to assess the effectiveness of each component, as shown in Table~\ref{tab:ablation} for both density-map-based (SASNet) and point-regression-based (P2PNet) models. For density-map paradigms, the baseline combination with $\mathcal{L}_{\text{hinge}}$ yields weaker attacks (MR $\approx45\%$). Adding $\mathcal{L}_{\text{cam}}$ or $\mathcal{L}_{\text{freq}}$ individually drives MR to approximately $60\%$, demonstrating these components are critical for disrupting density predictions. The full ensemble achieves the highest MR ($60.89-60.90\%$) though at a fair visual quality (PSNR: $17.46-17.47$ dB). For point-regression, the baseline achieves $45.15\%$ MR. Adding $\mathcal{L}_{\text{cam}}$ yields the best overall trade-off (MR: $45.61\%$, PSNR: $19.10$ dB). Unlike density maps, incorporating frequency or smoothness constraints consistently reduces MR, suggesting these priors interfere with optimization against point-regression outputs. The ablation confirms that all loss components contribute, but their impact is paradigm-dependent: frequency-domain constraints are essential for density maps, while attention-guided perturbations benefit point-regression models while better preserving imperceptibility.
\section{Conclusion} \label{sec:conc}
In this work, we proposed the first cross-paradigm digital adversarial attack framework for localized crowd counting, demonstrating universal vulnerabilities across state-of-the-art density map-based and point regression-based design paradigms. Our method achieves the optimal effectiveness-quality balance, compromising counting accuracy while maintaining competitive visual quality. The alarming transferability rate of our proposed attacks with ratios up to $1.69$ across seven models, revealing fundamental security gaps in current crowd counting systems. In future work, we want to examine the extent this phenomenon extend to physical attack. This work establishes new adversarial benchmarks and underscores the critical need for paradigm-agnostic defenses in safety-critical crowd analysis applications. 
\FloatBarrier
{
    \small
    \bibliographystyle{ieeenat_fullname}
    \bibliography{references}
}

\end{document}